\documentclass[10pt,twocolumn,letterpaper]{article}

\usepackage[pagenumbers]{iccv} 

\usepackage{multirow}
\usepackage[normalem]{ulem}
\useunder{\uline}{\ul}{}
\usepackage{adjustbox} 
\usepackage{amssymb} 
\usepackage{float} 
\usepackage{multirow}
\usepackage{array}
\usepackage{tabularx}

\usepackage{graphicx}
\usepackage{subcaption} 
\usepackage{float}      
\usepackage{mathrsfs} 

\definecolor{iccvblue}{rgb}{0.21,0.49,0.74}
\usepackage[pagebackref,breaklinks,colorlinks,allcolors=iccvblue]{hyperref}

\def\paperID{11284} 
\def\confName{ICCV}
\def\confYear{2025}

\title{Text-IRSTD: Leveraging Semantic Text to Promote Infrared Small Target Detection in Complex Scenes}

\author{
Feng Huang, Shuyuan Zheng, Zhaobing Qiu, Huanxian Liu, Huanxin Bai, Liqiong Chen\\
}

\begin{document}
\maketitle
\begin{abstract}
Infrared small target detection is currently a hot and challenging task in computer vision. Existing methods usually focus on mining visual features of targets, which struggles to cope with complex and diverse detection scenarios. The main reason is that infrared small targets have limited image information on their own, thus relying only on visual features fails to discriminate targets and interferences, leading to lower detection performance. To address this issue, we introduce a novel approach leveraging semantic text to guide infrared small target detection, called Text-IRSTD. It innovatively expands classical IRSTD to text-guided IRSTD, providing a new research idea. On the one hand, we devise a novel fuzzy semantic text prompt to accommodate ambiguous target categories. On the other hand, we propose a progressive cross-modal semantic interaction decoder (PCSID) to facilitate information fusion between texts and images. In addition, we construct a new benchmark consisting of 2,755 infrared images of different scenarios with fuzzy semantic textual annotations, called FZDT. Extensive experimental results demonstrate that our method achieves better detection performance and target contour recovery than the state-of-the-art methods. Moreover, proposed Text-IRSTD shows strong generalization and wide application prospects in unseen detection scenarios. The dataset and code will be publicly released after acceptance of this paper.
\end{abstract}


\section{Introduction}
\label{sec:intro}
Infrared small target detection (IRSTD) has become an important and challenging issue in computer vision, which is widely used in both civil and military applications \cite{Lin2024LCSRNet,liu2024infrared,tong2024sttrans}. However, due to the long range detection needs, imaged targets usually occupy only a few pixels and lack texture details \cite{ying2023mapping,zhang2022isnet,wang2024afe}. Therefore, how to accurately detect infrared (IR) small targets from complex backgrounds remains an open issue.
\begin{figure}[h]
    \centering
    \includegraphics[width=0.5\textwidth]{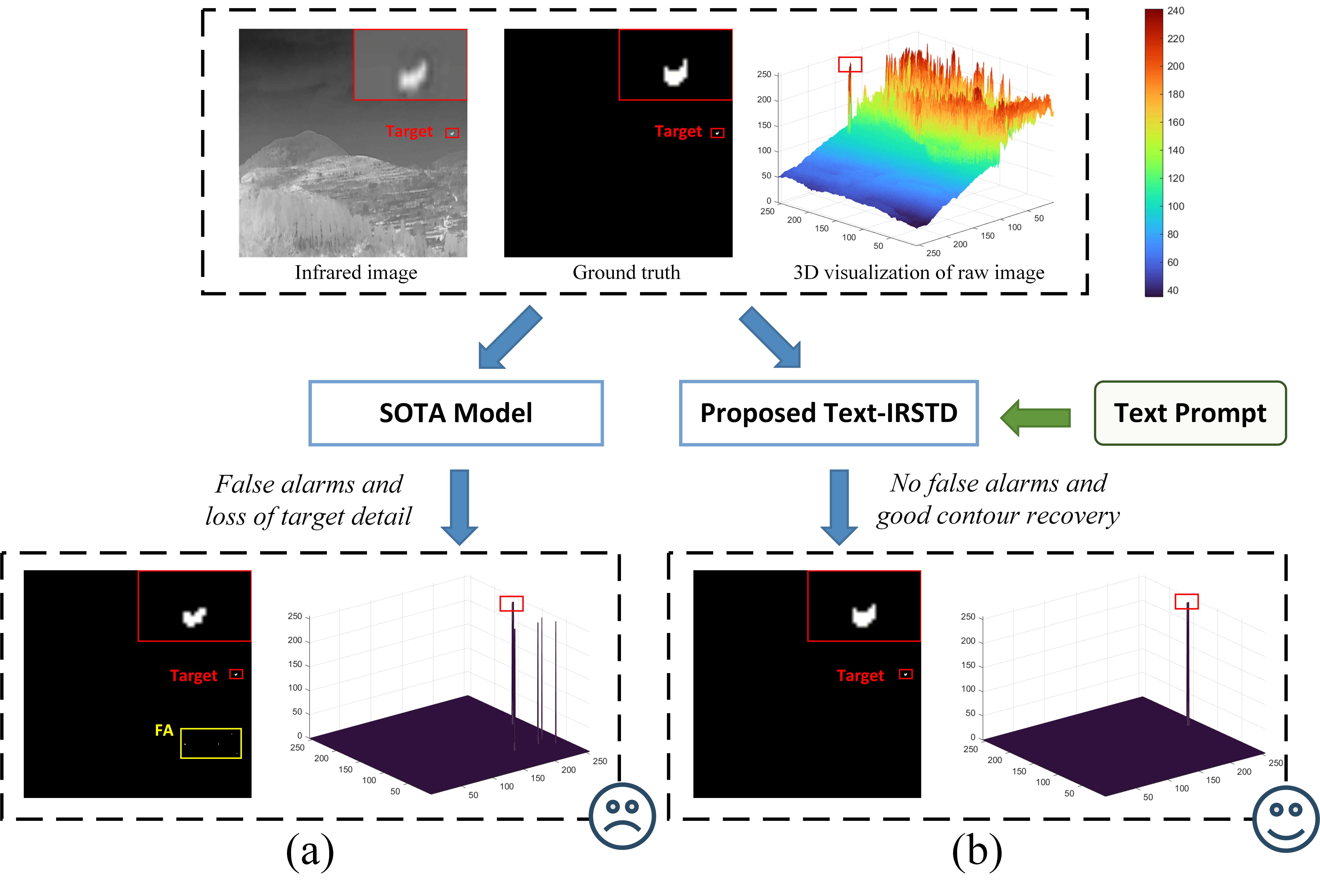}
    \caption{Visual comparison between the SOTA model and proposed Text-IRSTD. (a) Detection results of the SOTA model \cite{yuan2024sctransnet}, which relies only on visual features. (b) Detection results of proposed Text-IRSTD, which uses both textual and visual features.}
    \label{fig:Intro}
\end{figure}

In the early days, many traditional methods emerged to improve detection accuracy. They mainly consist of filter-based methods \cite{xin2017dual,Li2023GFilter}, local saliency-based methods \cite{xu2023infrared,zhang2023infrared,qiu2024RUMFR} and sparse representation-based methods \cite{NTFRA2022,Luo2022IMMN,Yan2023SCPNTLA}. However, reliance on artificial feature representations prevents them from further mining the deep semantic information of real targets. In contrast, deep learning (DL)-based methods autonomously learn multi-layered features in a data-driven manner and achieve significant performance improvements. For example, Dai et al. \cite{dai2021ACM} propose asymmetric contextual modulation (ACM) module to better interact with high-level semantics and low-level detail features. For higher feature utilization, Zhang et al. \cite{zhang2023AGPC} propose attention-guided pyramid context network (AGPCNet), which mines the correlations between feature map patches. In addition, researchers have also attempted to fuse prior information for higher detection accuracy \cite{Hou2021RISTDnet,yu2022pay}.

However, limited by scarce image features, existing methods fail to perform satisfactory in complex scenes. As shown in Fig. \ref{fig:Intro}, the SOTA model still suffers from many false detections due to limited ability to discriminate between targets and interference. Recently, with the development of large-scale visual language models \cite{radford2021clip}, semantic text-image-based models \cite{li2022glip,wu2023cora} have made impressive progress in generic target detection. However, applying them directly to infrared small target detection works poorly. The main reason is that ambiguous target categories in infrared images invalidate existing data annotation and feature fusion methods. 

To address above challenges, this paper proposes a novel approach leveraging semantic text to guide infrared small target detection, called Text-IRSTD. It innovatively expands classical IRSTD to text-guided IRSTD, providing a new idea for IRSTD. As shown in Fig. \ref{fig:Intro}, by utilizing both text and image information, proposed Text-IRSTD can further suppress clutter interference and improve detection performance. On the one hand, we devise a novel fuzzy semantic text prompt to accommodate ambiguous target categories. This prompt provides more text-image semantic associations that contribute to clutter suppression and target contour recovery. On the other hand, we propose a progressive cross-modal semantic interaction decoder (PCSID) that fuses both text and image information to further improve the detection performance. PCSID consists of two core components: text-guided feature aggregation (TGFA) block and text-guided semantic interaction (TGSI) block. Specifically, TGFA first utilizes text information to guide cross-layer visual feature aggregation for key feature extraction. Where key features (e.g., contrast \cite{xu2023infrared}) refer to those features that are more important than others (e.g., intensity) for IRSTD. Then, TGSI block simultaneously performs feature modulation from two different perspectives, thus realizing a more refined text-image information interaction. In addition, we construct a new benchmark consisting of 2,755 infrared images of different scenarios with fuzzy semantic textual annotations, called FZDT. Extensive experiments on multiple public datasets demonstrate that our approach achieves better detection performance and generalization than representative state-of-the-art algorithms. 





\section{Related Work}
In this section, we first briefly review existing methods for IRSTD. Then, an introduction to text-image models in other domains is presented.

\label{sec:RW}
\subsection{IRSTD Methods}
In the early days, many traditional methods emerged to improve detection accuracy. They mainly consist of filter-based methods \cite{xin2017dual,Li2023GFilter}, local saliency-based methods \cite{xu2023infrared,zhang2023infrared,qiu2024RUMFR} and sparse representation-based methods \cite{NTFRA2022,Luo2022IMMN,Yan2023SCPNTLA}. However, reliance on artificial feature representations prevents them from further mining the deep semantic information of real targets.  In contrast, deep learning (DL)-based methods autonomously learn multi-layered features in a data-driven manner and have made some impressive progresses recently. To mitigate feature loss of IR small targets, several researchers focus on improving the cross-layer feature fusion approach. For example, Dai et al. \cite{dai2021ACM} propose an asymmetric contextual modulation (ACM) module to better interact with high-level semantics and low-level detail features. For higher feature utilization, Zhang et al. \cite{zhang2023AGPC} propose the attention-guided pyramid context network (AGPCNet), which mines the correlations between feature map patches. In order to interact cross-layer features incrementally, Li et al. \cite{li2022DNA} propose dense nested attention network (DNA-Net). Several researchers also attempted to improve the model structure. For example, Wu et al. \cite{wu2022UIU} embed a smaller U-Net into the U-Net backbone to learn target features from multiple levels and scales. In addition, Hou et al. \cite{Hou2021RISTDnet} try to introduce prior information and design a robust infrared small target detection network (RISTDnet) combining artificial features and neural networks.

Existing methods have made many impressive progress in IRSTD. However, the visual features of IR small targets are quite limited, which renders their performance unsatisfactory in complex and diverse detection scenarios. In contrast, we explore a new research idea to introduce text semantic features into IRSTD and devise a novel text-image fusion detection method. Utilizing added textual information, proposed Text-IRSTD can further suppress background interference and improve detection performance.

\subsection{Text-Image Models}
With the advancement of deep learning techniques and the support of large datasets, text-image models has achieved success. In particular, CLIP \cite{radford2021clip} utilizes 400 million matched pairs datasets and contrast learning techniques to demonstrate powerful performance and wide application prospects in multiple domains and tasks. CLIP significantly improves text-image alignment and avoids the reliance on explicit labeling. With the support of CLIP models, text-driven generic target detection has also developed rapidly. 

For better detection performance, RegionCLIP \cite{zhong2022regionclip} leverages CLIP to match image regions with template captions, achieving fine-grained alignment between image and text features. ViLD \cite{gu2021ViLD} designs a training method based on knowledge distillation, which distills the knowledge from a pretrained open-vocabulary image classification model into a two-stage detector. CORA \cite{wu2023cora} mitigates the distribution gap between overall image features and regional features when using CLIP directly for detector training by introducing learnable prompts.

Although existing text-image models for generic target detection are illuminating, they are difficult to apply to IRSTD tasks. The reason is that it is difficult to establish appropriate text-image semantic associations and interactions for IR small targets with ambiguous categories. Therefore, it is important and challenging to investigate text-image detection models dedicated to IRSTD tasks.

\section{Fuzzy Semantic Text Prompt For IRSTD }
\label{sec:TI-IRSTD}
\begin{figure}[h]
    \centering
    \includegraphics[width=0.5\textwidth]{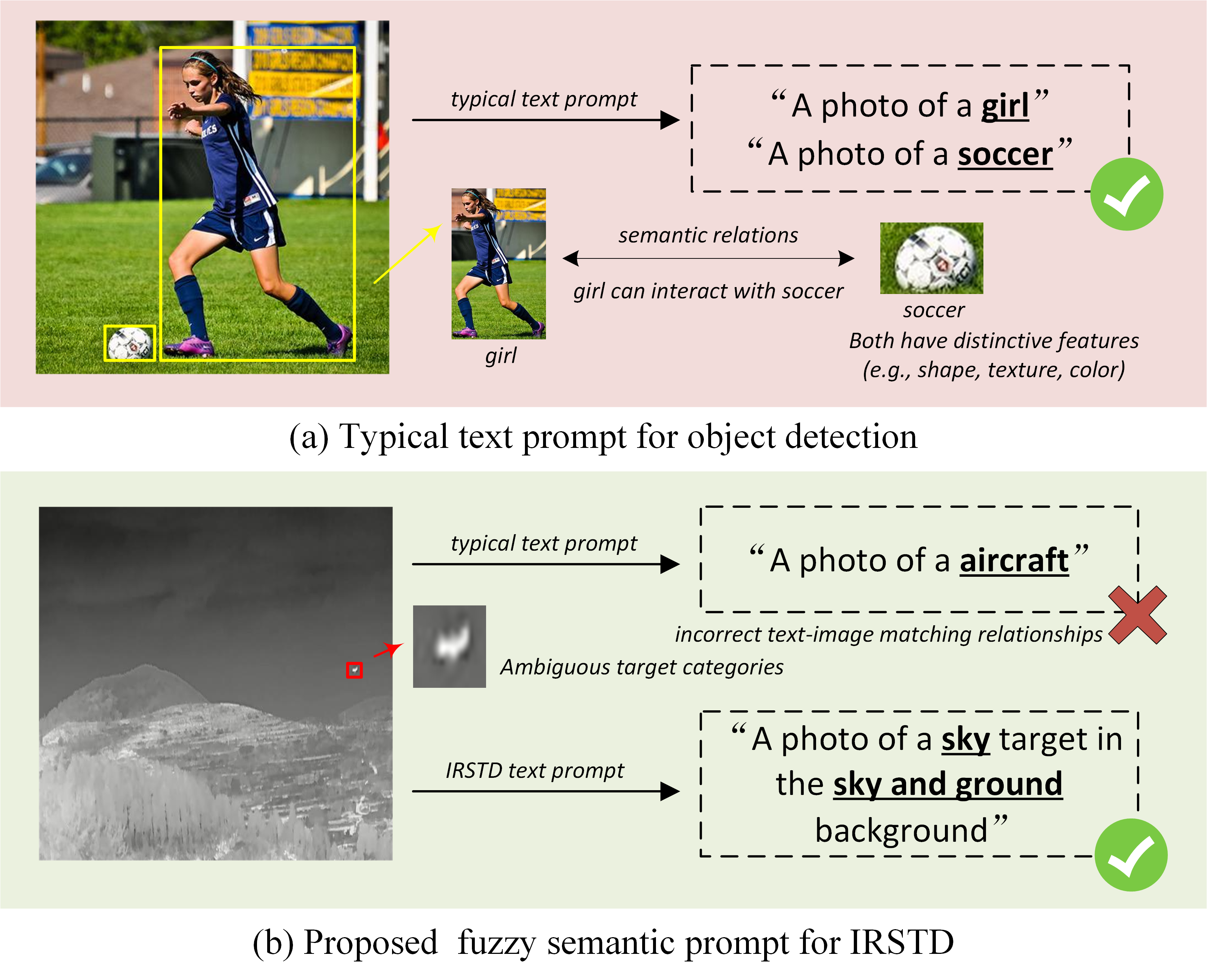}
    \caption{(a) Typical semantic text prompt for generic target detection. (b) Proposed fuzzy semantic text prompt for IRSTD.}
    \label{fig:dataset}
\end{figure}

Semantic text prompts aim to establish a proper matching relationship between text and images, which is crucial for cross-modal target detection. In generic target detection \cite{zhong2022regionclip,gu2021ViLD}, researchers usually directly describe specific categories of targets, such as “$\texttt{A photo of a girl}$” and “$\texttt{A photo of a soccer}$” in Fig. \ref{fig:dataset}(a). In this way, it is possible to focus not only on the visual features of a single target, but also to further explore semantic relations between multiple targets, e.g. a small round target near a girl is more likely to be a soccer. However, IR small targets typically occupy only a few pixels and lack visual features such as texture. As shown in Fig. \ref{fig:dataset}(b), real targets are usually presented as bright spots that cannot be further identified as a specific category such as an aircraft or a bird. Therefore, using existing text prompts to describe ambiguous IR small target categories will lead to incorrect text-image matching relationships, impairing detection accuracy and model generalizability. 

To address this issue, we propose a novel fuzzy semantic text prompt for IRSTD. Specifically, we describe one IR image as “\texttt{A photo of a [Interested Region] target in the [Scene] background}”, where $\texttt{[Interested Region]}$ denotes the region where targets need to be detected, $\textit{e.g.}$, sky target and ground target, and $\texttt{[Scene]}$ denotes the detection scene, $\textit{e.g.}$, sky-ground background, ocean-sky background. For example, we describe the IR image Fig. \ref{fig:dataset}(b) as: “\texttt{A photo of a sky target in the sky and ground background}”, which not only establishes correct and robust matching relationships, but also provides more semantic associations. In this way, the model can focus on more shared visual features of sky targets, such as shape contours, which helps target shape recovery and detection. Moreover, as shown in Fig. \ref{fig:Intro}, more semantic associations are established and utilized to further suppress complex ground interference and improve detection accuracy.
\section{Method}
\label{sec:Method}

In this section, we first present the overall architecture of proposed Text-IRSTD. Then, Sections \ref{sec:Feature Enhancement Module} and \ref{sec:text-semantic-guidance} introduce two core components: TGFA block for aggregating key visual features and TGSI block for facilitating text-image information interaction, respectively.

\subsection{Overall Architecture}
\begin{figure*}[h]
    \centering
    \includegraphics[width=1\textwidth]{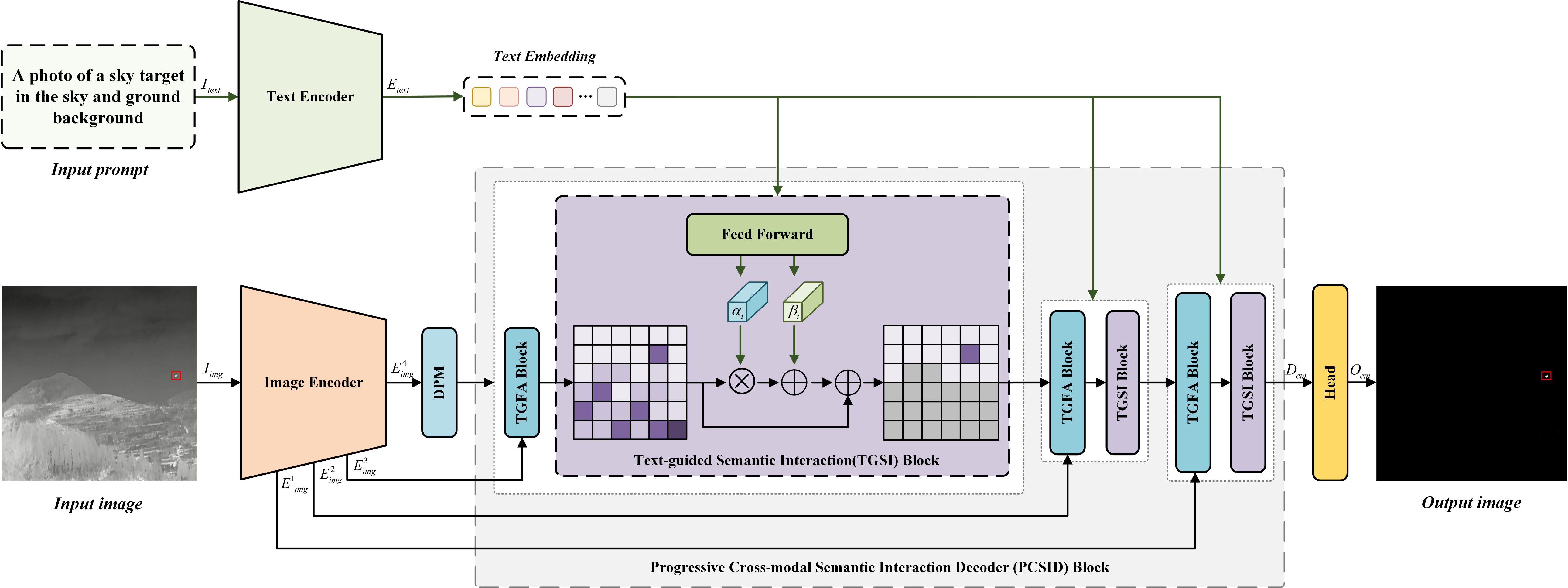}
    \caption{Overview of the proposed Text-IRSTD. It consists of three main components: text encoder, image encoder, and cross-modal decoder PCSID with their outputs represented as $E_{text}$, $E_{img}^{(i)}$, and $D_{cm}$, respectively. Where DPM denotes detail perception module, TGFA Block denotes text-guided feature aggregation block and TGSI Block denotes text-guided semantic interaction block.}
    \label{fig:overall}
\end{figure*}
Fig. \ref{fig:overall} illustrates the overall structure of Text-IRSTD. First, one infrared image $I_{img}$ and the corresponding text prompt ${I}_{text}$ are fed into the image encoder $\mathcal{F}_{i}$ and text encoder $\mathcal{F}_{t}$, respectively, as follows:
\begin{equation}
\begin{aligned}
    E_{img} &= \mathcal{F}_{i}(I_{img}),  \\ 
    E_{text} &= \mathcal{F}_{t}({I}_{text}),
\end{aligned}
\end{equation}
where $E_{img}$ and $E_{text}$ denote the image and text encoder outputs respectively. Note that $\mathcal{F}_{i}$ consists of four groups of ResNeSt \cite{zhang2022resnest} blocks, while $\mathcal{F}_{t}$ denote the pre-trained language model CLIP with frozen weights.

Then, $E_{img}$ and $E_{text}$ are fed into PCSID for text-image feature interaction and enhancement. The result of cross-modal interaction $D_{cm}$ can be expressed as:
\begin{equation}
    D_{cm}= \text{PCSID}(E_{img}, E_{text}),
\end{equation}
where PCSID consists of three groups of TGFA blocks and TGSI blocks in series. TGFA and TGSI will be introduced in Sections \ref{sec:Feature Enhancement Module} and \ref{sec:text-semantic-guidance}, respectively. Note that the fourth layer output $E_{\text{img}}^{(4)}$ of $\mathcal{F}_{i}$ is enhanced by detail perception module (DPM) and then fed to PCSID. 

Finally, we apply a 1×1 convolution $f_{c}^{1}$ and sigmoid function to the decoder outputs $D_{cm}$. Thus, the cross-modal detection result $O_{cm}$ can be obtained as follows:
 \begin{align}
    O_{cm} &= \text{Sigmoid}(f_{c}^{1}(D_{cm})).
\end{align}

\subsection{Text-guided Feature Aggregation Block}\label{sec:Feature Enhancement Module}
\begin{figure}[h]
    \centering
    \includegraphics[width=0.48\textwidth]{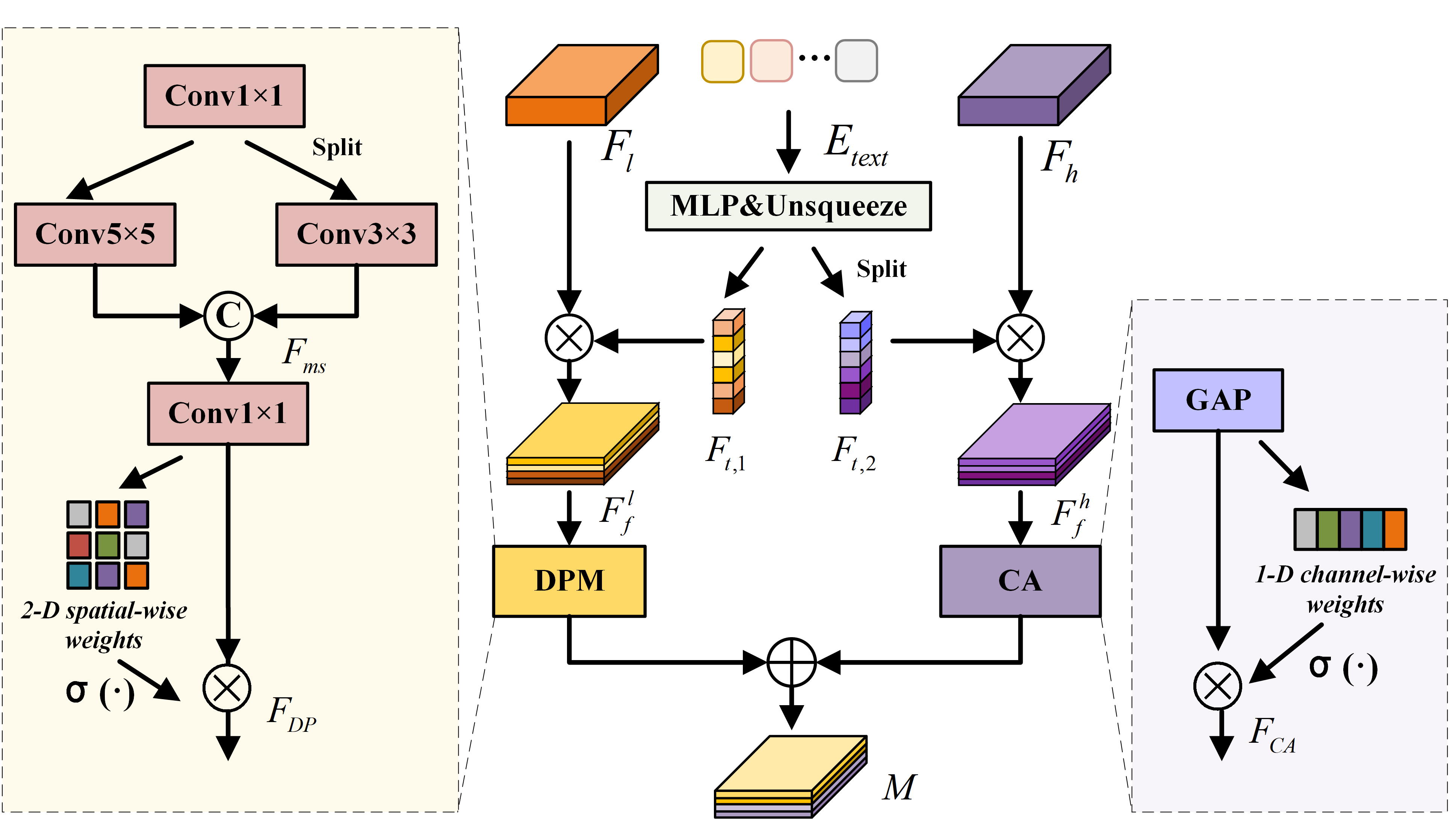}
    \caption{Structure of TGFA block,where DPM denotes detail perception module and CA denotes channel attention.}
    \label{fig:CLFA}
\end{figure}

Effective aggregation of visual features is crucial for detection models. Limited by scarce visual information, existing methods struggle to capture key features from high-level and low-level feature maps. To this end, we propose the novel text-guided cross-layer feature aggregation block TGFA (shown in Fig. \ref{fig:CLFA}). TGFA transforms text features $E_{text}$ into modulation information through MLP, guiding the model to capture key features that are easily neglected in complex scenes. The details are as follows:
\begin{equation}
F_{t,1}, F_{t,2} = \text{Spilt}(\text{MLP}(E_{text})),
\end{equation}
\begin{equation}
F_{f}^{l}=F_l \otimes F_{t,1}, F_{f}^{h}=F_h \otimes F_{t,2},
\end{equation}
where $F_{f}^{l}$ and $F_{f}^{h}$ denote the  fusion features after text modulation, $\text{Spit}(\cdot)$ denotes splitting the feature into two equal groups along the channel dimension, $\otimes$ denotes element-wise multiplication.

Low-level features contain rich target locations and details, but are easily lost at deep layers because small IR targets occupy only a few pixels. Therefore, we devise the detail perception module DPM as shown in Fig. \ref{fig:CLFA}. Specifically, for a given low-level fusion feature $F_{f}^{l}$, a $1\times 1$ convolution is first applied to expand the channel dimension by a factor of $\mu$, and then divide the resulting feature map into two groups $F_{l,1} \in \mathbb{R}^{\frac{\mu C}{2} \times H \times W}$ and $F_{l,2} \in \mathbb{R}^{\frac{\mu C}{2} \times H \times W}$ along the channel dimension.

Then, we use $3\times 3$ and $5\times 5$ depth-wise convolutions to capture important local features such as contrast at different scales and concatenate these two groups of features along the channel dimension. The multi-scale detailed feature mapping $F_{ms}$ is obtained from the following equation:
\begin{equation}
F_{l,1}, F_{l,2} = \text{Split}(f_c^1(F_{f}^{l})),
\end{equation}
\begin{equation}
F_{ms} = f_{c}^{1} \left( \text{Contact}\left( f_{dwc}^{3}(F_{l,1}), f_{dwc}^{5}(F_{l,2}) \right) \right),
\end{equation}
where $\text{Contact}(\cdot)$ denotes concatenating the two groups of features along the channel dimension,$f_{c}^{1}$ denotes 1 × 1 convolution, $f_{dwc}^{3}$ and $f_{dwc}^{5}$ denote $3\times 3$ and $5\times 5$ depth-wise convolutions. 

After obtaining the multi-scale detailed feature mapping $F_{ms}$, we use point-wise convolution to aggregate the contextual features at each spatial location. For simplicity, it can be described as:
\begin{equation}
    F_{DP} =\sigma\left(\text{PWConv}_2\left(\gamma\left(\text{PWConv}_1(F_{ms})\right)\right)\right) \otimes F_{ms},
\end{equation}
where $\text{PWConv}_1$ and $\text{PWConv}_2$ denote the point-wise convolution with kernel sizes of $\frac{C}{4} \times C \times 1 \times 1$ and $C \times \frac{C}{4} \times 1 \times 1$, $\sigma$ and $\gamma$ represent the sigmoid activation function and rectified linear unit.

Moreover, high-level features can provide more accurate semantic information, while adjusting the weights of different channels can adaptively enhance the representation of key information and suppress redundant features. Therefore, for a given high-level fusion features $F_{f}^{h}$, we first use global average pooling on each channel to obtain the global feature representation $z_c \in \mathbb{R}^{C \times 1 \times 1}$, which can be described as:
\begin{equation}
z_c = \frac{1}{H \times W} \sum_{i=1}^{H} \sum_{j=1}^{W} F_{f}^{h}[:,i,j] \label{GAP}.
\end{equation}

Then, a nonlinear mapping of this global information is performed through two fully connected layers to establish dependencies between the channels. 

Finally, the channel feature maps are normalized using the sigmoid function, and an element-wise multiplication is performed with the $F_{f}^{h}$, the output of CA is defined as:
\begin{equation}
F_{CA} = \sigma \left( Fc_2 \left( Fc_1 (z_c) \right) \right) \otimes F_{f}^{h},
\end{equation}
where $Fc_1 \in \mathbb{R}^{\frac{C}{r} \times C}$ and $Fc_2 \in \mathbb{R}^{C \times \frac{C}{r}}$ represent the two fully connected operations, $r$ denotes the channel reduction rate.

\subsection{Text-guided Semantic Interaction Block}\label{sec:text-semantic-guidance}
Due to the limited visual information, existing models struggle to effectively suppress complex interference and recover target contours. Therefore, we introduce semantic text prompts and propose TGSI block, which achieves better detection performance through text-image information interaction. Figure \ref{fig:overall} shows the detailed structure of TGSI block. First, we take the image feature $M$ and the text feature $E_{text}$ as inputs, where $E_{text}$ is obtained by feeding the text prompt $T_{text}$ into the CLIP. Notably, we freeze the weights of the text encoder to preserve the original capabilities of the pre-trained model.

Furthermore, we design a Feed-forward Network (FFN) to further explore the deep semantic information embedded in the text features. This helps the model capture more fine-grained semantic associations between the text, the target, and the regions of interest. Specifically, the text feature $E_{text}$ is mapped into two learnable semantic parameters $\alpha_t$ and $\beta_t$ after going through the FFN, which can be defined as:
\begin{equation}
\alpha_t,\ \beta_t = \text{FFN}(E_{text}).
\end{equation}

Finally, the semantic parameters $\alpha_t$ and $\beta_t$ modulate the image feature $M$ from the perspective of gain control and offset adjustment, respectively. Specifically, the gain parameter $\alpha_t$ directs the network to pay more attention to the regions of interest of the the text prompt and reduces false alarms in other regions. The offset parameter $\beta_t$ provides finer tuning and helps in the recovery of target contour details. The fused semantic features $\hat{M}$ are obtained by:
\begin{equation}
\hat{M} = (1 + \alpha_t) \otimes M + \beta_t.
\end{equation}

\section{Experiments}
\label{sec:Experiments}
\begin{table*}[h]
\caption{Quantitative results of different methods. Results for the metrics of IoU(\%), $\text{P}_d$(\%), $\text{F}_a$($10^{-6}$) and $\text{F}_1$(\%) are presented. The best values are highlighted with \textbf{bold}, the second best values are {\ul underlined}.}
\vspace{-2mm}
\centering
\renewcommand{\arraystretch}{1}
\begin{tabularx}{\textwidth}{c|>{\centering\arraybackslash}X>{\centering\arraybackslash}X>{\centering\arraybackslash}X>{\centering\arraybackslash}X|>{\centering\arraybackslash}X>{\centering\arraybackslash}X>{\centering\arraybackslash}X>{\centering\arraybackslash}X}
\hline
\multirow{2}{*}{Method} & \multicolumn{4}{c|}{NUDT-SIRST}                                   & \multicolumn{4}{c}{IRSTD-1k}                                      \\ \cline{2-9} 
                        & IoU           & $\text{P}_d$             & $\text{F}_a$             & $\text{F}_1$             & IoU           & $\text{P}_d$             & $\text{F}_a$             & $\text{F}_1$             \\ \hline
IPI \cite{gao2013IPI}                     & 38.15          & 90.78          & 369.9          & 61.52          & 30.95          & 78.91          & 180.4          & 55.15          \\
PSTNN \cite{zhang2019PSTNN}                   & 17.21          & 54.47          & 81.62          & 29.45          & 25.58          & 62.96          & 78.91          & 41.35          \\
FKRW \cite{qin2019fkrw}                    & 25.38          & 60.26          & 107.5         & 38.54          & 15.89          & 54.54          & 26.75          & 27.08          \\
GSWLCM \cite{qiu2022GSWLCM}                  & 4.909          & 65.00          & 15.71          & 9.301          & 3.068           & 59.25          & 26.94          & 6.045          \\
RUMFR \cite{qiu2024RUMFR}                   & 26.08          & 67.89          & 96.42          & 39.25          & 10.26          & 53.87          & 25.29          & 18.59          \\ \hline
ACMNet \cite{dai2021ACM}                  & 68.14          & 96.57          & 17.38          & 83.41          & 61.14          & 87.88          & 41.79          & 74.17          \\
ALCNet \cite{dai2021ALC}                  & 72.56          & 96.05          & 7.572          & 86.78          & 58.27          & 89.90          & 44.69          & 72.92          \\
AGPCNet \cite{zhang2023AGPC}                 & 85.97          & 97.63          & 7.228          & 93.51          & 62.51          & 91.58          & 20.97          & 77.68          \\
DNANet \cite{li2022DNA}                  & 94.70           & 98.68    & 3.212          & 97.31          & 64.51          & 90.23          & 18.27          & 78.77          \\
UIUNet \cite{wu2022UIU}                  & 89.15          & 97.11          & 5.506          & 94.23          & 64.32          & 90.24          & 28.33          & {\ul79.02}          \\
RDIAN \cite{sun2023RDIAN}                   & 86.08          & 98.15          & 6.711          & 92.36          & 61.86          & 88.56          & 41.44          & 76.27          \\
DMFNet \cite{guo2024dmfnet}                  & 87.67          & 98.42          & 2.294          & 93.30          & 64.45          & 90.14          &  18.24    & 78.52          \\
SCTrans \cite{yuan2024sctransnet}                 & 94.83          & 98.42          &  2.065    & 97.19          & 62.95          & {\ul 92.24}    & 23.40          & 78.04          \\
Text-IRSTD (w/o text)      & {\ul 95.25}    & {\ul 98.94}    & {\ul1.663}          & {\ul 97.37}    & {\ul 65.50}    & 91.56          & {\ul16.13}          & 78.73 \\
Text-IRSTD (Ours)          & \textbf{95.84} & \textbf{99.73} & \textbf{1.032} & \textbf{97.95} & \textbf{69.57} & \textbf{92.59} & \textbf{14.97} & \textbf{79.24}    \\ \hline
\end{tabularx}
\label{table:results}
\end{table*}

\subsection{Datasets and Evaluation Metrics}
\textbf{{Datasets:}} All experiments are conducted on proposed FZDT dataset, which contains three public image datasets NUAA-SIRST \cite{dai2021ACM}, NUDT-SIRST \cite{li2022DNA}, IRSTD-1k \cite{zhang2022isnet}, and corresponding fuzzy semantic text prompts. Following existing works \cite{yang2024eflnet}, each dataset is divided into training set, validation set and test set in the ratio of $6:2:2$.

\textbf{{Evaluation Metrics:}} For comprehensive algorithmic comparison, the evaluation metrics include intersection over union (IoU), false alarm rate ($\text{F}_a$), F-measure ($\text{F}_1$), and probability of detection ($\text{P}_d$) \cite{yuan2024sctransnet}. 

\subsection{Implementation Details}
To evaluate the efficacy of Text-IRSTD, we use the common CLIP-ViT-B/32 \cite{radford2021clip} as the text encoder. Our network is implemented with Pytorch on a single NVIDIA GeForce RTX 3090 GPU. We adopt the AdamW optimizer with a learning rate of 0.0001 and a weight decay of 0.05. Every images are normalized and randomly cropped to $256 \times 256$. The batch size and epoch are set as 8 and 500, respectively. 

The comparison algorithms for experiments include eight state-of-the-art DL-based methods: ACMNet \cite{dai2021ACM} , ALCNet \cite{dai2021ALC} , AGPCNet \cite{zhang2023AGPC} , DNANet \cite{li2022DNA} , UIUNet \cite{wu2022UIU} , RDIAN \cite{sun2023RDIAN} , DMFNet \cite{guo2024dmfnet} and SCTrans \cite{yuan2024sctransnet}, and five recent traditional methods: IPI \cite{gao2013IPI} , PSTNN \cite{zhang2019PSTNN} , FKRW \cite{qin2019fkrw} , GSWLCM \cite{qiu2022GSWLCM} and RUMFR \cite{qiu2024RUMFR}. To further illustrate the advancement of our method, Text-IRSTD without text prompts is also selected as the baseline method.

\subsection{Quantitative Results}
The quantitative results of different methods are presented in Tab. \ref{table:results}. Overall, our Text-IRSTD achieves the best performance across all evaluation metrics. For example, the $\text{F}_a$ of our method achieves as low as 1.032 $\times 10^{-6}$ on NUDT-SIRST, while also reaching the highest $\text{P}_d$ of 99.73\%, demonstrating our superior capability in suppressing complex interference. Additionally, we obtain the best IoU of 95.84\%, which is 1.01\% higher than the second best, indicating better target contour recovery capability. Traditional methods have lower metrics and exhibit poor performance. DL-based methods can further mine deep semantic features in a data-driven manner, and thus have significant improvements in all metrics. However, the limited visual information of the target makes it difficult to further improve their detection accuracy in complex scenes. For instance, these SOTA DL-based methods achieve a maximum IoU of only 64.51\% on the IRSTD-1k, which is difficult to satisfy higher detection requirements. By incorporating text information and designing cross-modal information interaction methods, the proposed Text-IRSTD achieves up to 69.57\% performance improvement in IoU.

It is worth noting that our method is able to produce satisfactory results even when using the default text input (‘w/o text’) in the testing stage. This phenomenon indicates that our method does not rely completely on detailed text prompts, and the inherent advantages of the model design contribute to the performance of IRSTD.


\subsection{Qualitative Results}
\begin{figure*}[h]
    \centering
    \includegraphics[width=1\textwidth]{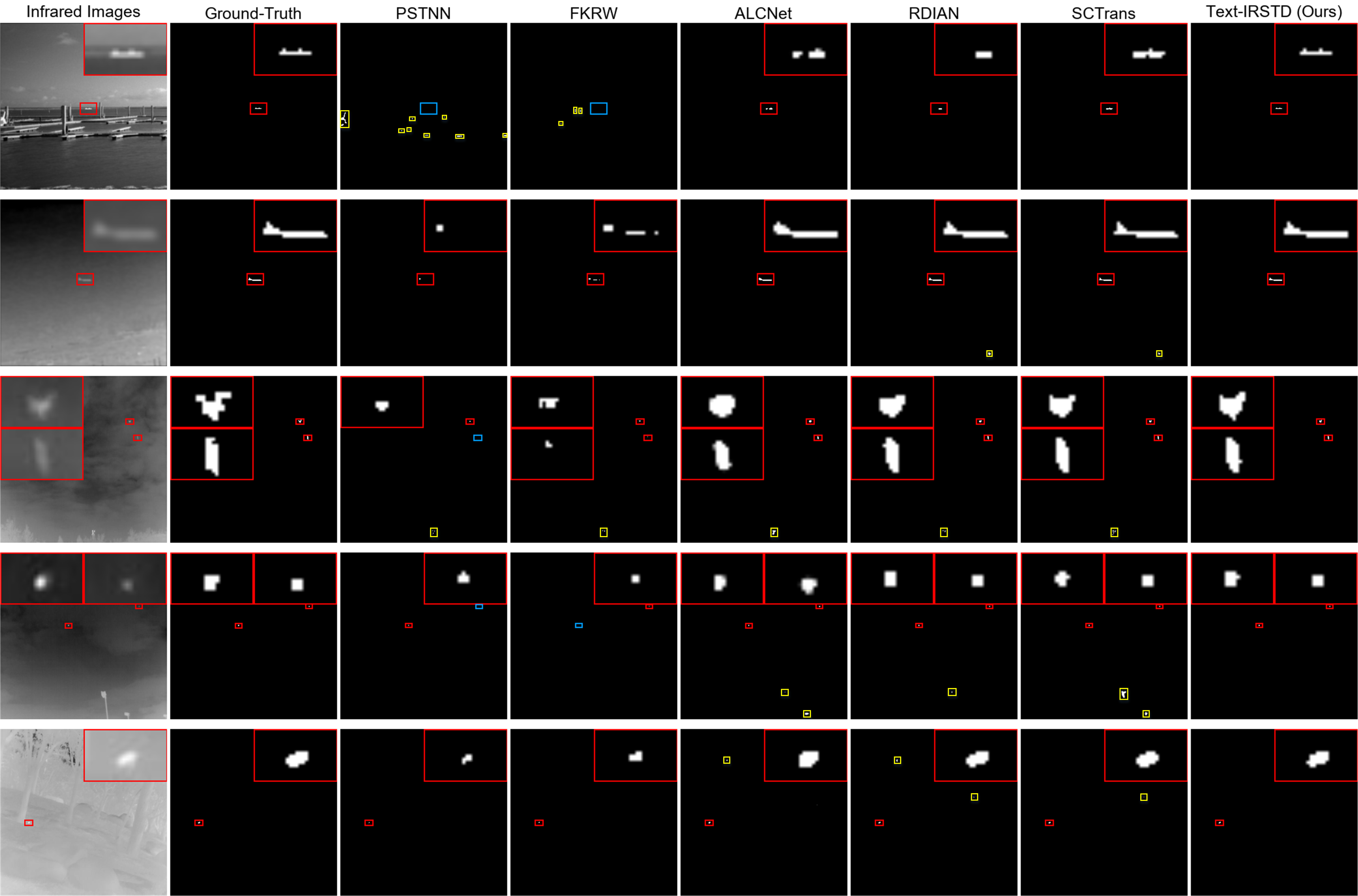}
    \caption{Visual results of different IRSTD methods. The enlarged view of the target is shown in image corners. The red, blue, and yellow boxes represent correctly detected targets, missed targets, and false detections, respectively.}
    \label{fig:visual}
\end{figure*}

\begin{figure*}[h]
    \includegraphics[width=1\textwidth]{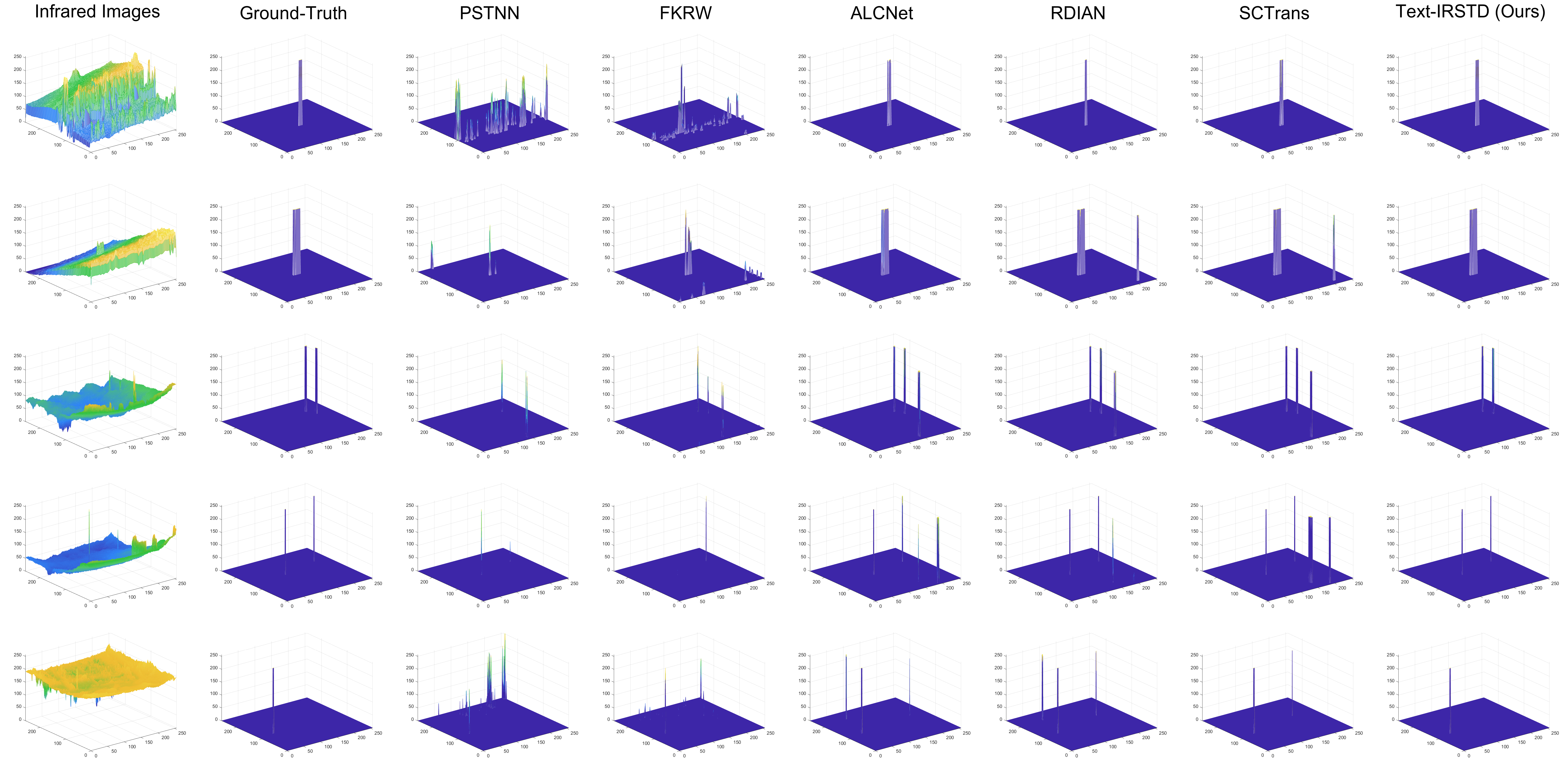}
    \caption{3D visualization results corresponding to those in Fig. \ref{fig:visual}. Different colors represent distinct grayscale values, gradually increasing from blue to yellow.}
    \centering
    \label{fig:3D}
\end{figure*}

We visualize the detection results of six representative algorithms, as shown in Fig. \ref{fig:visual} and Fig. \ref{fig:3D}. It can be seen that traditional algorithms (e.g., PSTNN) have a large number of missed detections and false alarms due to their limited artificial feature representation capabilities. DL-based methods have made significant progress in background suppression by mining high-level semantic features. However, they all showed false alarms in the third row of images in Fig. \ref{fig:visual}. The reason is that the limited visual information of infrared images makes it difficult to further suppress interference and improve detection performance.

In contrast, proposed Text-IRSTD accurately detects targets in every scene and suppresses almost all background interference. By introducing semantic text to expand the limited visual information, Text-IRSTD can further improve the detection accuracy in complex scenes. As shown in Fig. \ref{fig:visual}, thanks to the PCSID, we realize high-performance interactions across modal information, which can focus more on the region of interest to suppress complex interferences, while perceive and enhance the detail information to guide the accurate detection and contour recovery of targets.

\subsection{Performance in Unseen Scenarios}
In practical applications, models often need to face unseen detection scenarios. The detection performance in unseen scenarios can evaluate the practicality and generalization of models to some extent. To further validate the advancement of proposed methods, we compare the detection capabilities of different methods in unseen scenarios.

Specifically, we train different models using the NUDT-SIRST dataset and then test their detection performance using the unseen NUAA-SIRST dataset. Tab. \ref{table:Generalization} presents the detection results of different models on the unseen dataset. It can be seen that our method outperforms the state-of-the-art models in terms of IoU, $\text{P}_d$ and $\text{F}_a$, implying better detection accuracy and scene generalization.
\begin{table}[H]
\caption{Detection performance of different methods for the unseen dataset.}
\vspace{0mm}
\centering
\renewcommand{\arraystretch}{1.1}
\resizebox{1\linewidth}{!}{ 
\begin{tabularx}{\linewidth}{c|>{\centering\arraybackslash}X>{\centering\arraybackslash}X>{\centering\arraybackslash}X}
\hline
\multirow{2}{*}{Method} & \multicolumn{3}{c}{NUDT-SIRST (Train)} \\ 
                        & \multicolumn{3}{c}{NUAA-SIRST (Test)}  \\ \cline{2-4}
                        & IoU(\%)            & $\text{P}_d$(\%)               & $\text{F}_a$($10^{-6}$)               \\ \hline
DNANet \cite{li2022DNA}  & 35.80            & 76.14            & 141.3            \\
UIUNet \cite{wu2022UIU}  & {\ul 53.32}     & 80.73            & {\ul 35.15}             \\
RDIAN  \cite{sun2023RDIAN} & 39.35           & 77.98            & 43.68             \\
DMFNet \cite{guo2024dmfnet}  & 43.70           & 79.81            & 60.07            \\
SCTrans \cite{yuan2024sctransnet} & 50.97            & {\ul 83.48}      & 37.71            \\
Text-IRSTD (Ours)       & \textbf{54.51}  & \textbf{87.15}   & \textbf{29.86}    \\ \hline
\end{tabularx}
} 
\label{table:Generalization}
\end{table}

\subsection{Ablation Study}

\begin{table*}[h]  
\centering
\caption{Ablation study of different prompts.}
\vspace{0mm}
\renewcommand{\arraystretch}{1.1}
\resizebox{1\linewidth}{!}{
\begin{tabular}{c|c|ccc}
\hline
   Type & Prompt                                                                             & IoU(\%) & $\text{P}_d$(\%) & $\text{F}_a$($10^{-6}$)       \\ \hline
 Conventional prompts & A photo of a [class]                                                           & 62.22    & 90.58   & 30.36 \\ \hline
Variants of convention & A photo of a [class] in the [Scene] background
 &66.65     &89.89     &21.69  \\ \hline
 Variants of Ours & A [Interested Region] target in the [Scene] background  & {\ul69.01}   & {\ul91.68}   & {\ul18.86} \\ \hline
Ours & A photo of a [Interested Region] target in the [Scene] background
          & \textbf{69.57}    & \textbf{92.59}   & \textbf{14.97} \\ \hline
\end{tabular}}
\label{text prompt}
\end{table*}

\textbf{Impact of Text Prompt:} In order to verify the effectiveness of fuzzy text prompts, we designed three textual prompt variants for experiments and the results are shown in Tab. \ref{text prompt}. It can be seen that the use of $\texttt{[Interested Region] target}$ as a prompt to describe the IR small target helps to avoid the wrong text-image matching relationship caused by the use of explicit categories of prompts. Meanwhile, the description of $\texttt{[Scene] background}$ further improves the detection performance in complex scenes by establishing more appropriate semantic associations. More importantly, experiments demonstrate the possibility of improving IRSTD performance with text and provide an effective technical path for further research.

\textbf{Impact of PCSID:} To validate the effectiveness of each component in PCSID, we perform several ablation studies on the IRSTD-1k, and the results are shown in Tab. \ref{table:ablation}. The results of the experiment indicate that both TGSI block and TGFA block improve the performance of the baseline model, and when use together, the model achieve the best results. This further confirms the importance of their synergistic effect in enhancing the performance of small target detection.
\begin{table}[h]
\caption{Ablation study of the TGFA block and TGSI block.}
\vspace{0mm}
\centering
\renewcommand{\arraystretch}{1.1} 
\begin{tabularx}{\linewidth}{>{\centering\arraybackslash}m{1cm} >{\centering\arraybackslash}m{0.9cm} >{\centering\arraybackslash}m{0.9cm} >{\centering\arraybackslash}m{0.9cm} >{\centering\arraybackslash}m{0.9cm} >{\centering\arraybackslash}m{1cm}}
\hline
Baseline & TGFA & TGSI & IoU(\%) & $\text{P}_d$(\%) & $\text{F}_a$($10^{-6}$) \\ \hline
\checkmark & & & 61.67 & 89.22 & 57.97 \\
\checkmark & \checkmark & & {\ul67.75} & 91.87 & {\ul19.83} \\
\checkmark & & \checkmark &  66.19 & {\ul 91.92} &  26.39 \\
\checkmark & \checkmark & \checkmark & \textbf{69.57} & \textbf{92.59} & \textbf{14.97} \\ \hline
\end{tabularx}
\label{table:ablation}
\end{table}

\begin{table}[h]
\caption{Ablation study of the TGFA block.}
\vspace{0mm}
\centering
\renewcommand{\arraystretch}{1.1}
\begin{tabularx}{\linewidth}{>{\centering\arraybackslash}m{0.4\linewidth} >{\centering\arraybackslash}m{0.14\linewidth} >{\centering\arraybackslash}m{0.14\linewidth} >{\centering\arraybackslash}m{0.14\linewidth}}
\hline
Model               & IoU(\%)           & $\text{P}_d$(\%)             & $\text{F}_a$($10^{-6}$)            \\ \hline
TGFA w/o text guide                & {\ul 68.75 }         & {\ul 92.02}    & 23.68          \\
TGFA w/o CA                & 68.01    & 91.24          &  26.24    \\
TGFA w/o DPM  & 67.16          & 89.56          & {\ul17.95 }        \\ 
TGFA (Ours)                  & \textbf{69.57} & \textbf{92.59} & \textbf{14.97} \\ \hline
\end{tabularx}
\label{table:CLFA}
\end{table}

\begin{table}[h]
\caption{Ablation study of the TGSI block.}
\vspace{0mm}
\centering
\renewcommand{\arraystretch}{1.1}
\begin{tabularx}{\linewidth}{>{\centering\arraybackslash}m{0.3\linewidth} >{\centering\arraybackslash}m{0.17\linewidth} >{\centering\arraybackslash}m{0.17\linewidth} >{\centering\arraybackslash}m{0.17\linewidth}}
\hline
Model               & IoU(\%)           & $\text{P}_d$(\%)              & $\text{F}_a$($10^{-6}$)            \\ \hline
Concatenation         & 67.73          & 90.57          & 29.49          \\
TGSI w/o $\alpha_t$        & 66.65    & 90.13          &  35.85     \\
TGSI w/o $\beta_t$         & {\ul68.05 }         & {\ul 91.21}     &  {\ul16.22}         \\
TGSI (Ours)                 & \textbf{69.57} & \textbf{92.59} & \textbf{14.97} \\ \hline
\end{tabularx}
\label{table:settings}
\end{table}


\textbf{Impact of TGFA:} To demonstrate the effectiveness of TGFA block, we introduce three network variants for comparison, as shown in Tab. \ref{table:CLFA}. It can be seen that proposed TGFA block achieves the best performance. The reason is that text features as modulation information can optimize the complementarity and interaction between cross-layer features. Moreover, DPM and CA can better perceive and enhance important visual features, such as contrast, which facilitate visual feature fusion and subsequent cross-modal information interaction.

\textbf{Impact of TGSI Block:} We use four different text-image feature interaction methods to train the Text-IRSTD. As shown in Tab. \ref{table:settings}, simply concatenating text and image features along the channel dimension ignores semantic inconsistencies across modalities, leading to inadequate semantic alignment. Without gain parameter $\alpha_t$, the model cannot dynamically focus on the regions of interest indicated by the text prompt. Similarly, without offset parameter $\beta_t$, the model cannot make fine adjustments to the image features using the text features, thus limiting the effectiveness of target shape reconstruction. By combining both, the proposed TGSI Block achieves the best performance.

\section{Conclusion}
\label{sec:5_Conclusion}
In this paper, we extend the IRSTD task and propose a text-guided IRSTD framework to address the issue that current models struggle to further improve detection accuracy by relying only on limited visual information. On the one hand, we devise a novel fuzzy semantic text prompt to accommodate ambiguous target categories. This prompt provides more text-image semantic associations that contribute to clutter suppression and target contour recovery. On the other hand, we propose PCSID to facilitate the information interaction between text and images. Moreover, we have constructed the first text-image bimodal dataset, called FZDT, to provide the basis for subsequent text-guided IRSTD research. Extensive experimental results demonstrate that our method achieves better detection performance and target contour recovery than the state-of-the-art methods. More surprisingly, the proposed Text-IRSTD demonstrates strong generalization and wide application prospects in unseen detection scenarios.

{
    \small
    \bibliographystyle{ieeenat_fullname}
    \bibliography{main}
}

\end{document}